\documentclass[10pt,twocolumn,letterpaper]{article}

\usepackage{iccv}
\usepackage{times}
\usepackage{epsfig}
\usepackage{graphicx}
\usepackage{amsmath}
\usepackage{amssymb}
\usepackage{subcaption}
\usepackage{multirow}
\usepackage{diagbox}
\usepackage{tikz}
\usepackage{cellspace}
\usepackage{caption}
\usepackage{bbm}

\makeatletter
\def\blfootnote{\xdef\@thefnmark{}\@footnotetext}
\makeatother

\newcommand{\tikzcirclewhite}[2][black,fill=white]{\tikz[baseline=-0.5ex]\draw[#1,radius=#2] (0,0) circle ;}%
\newcommand{\tikzcircleblack}[2][black,fill=black]{\tikz[baseline=-0.5ex]\draw[#1,radius=#2] (0,0) circle ;}%

\newcommand*\halfcirc[2][2.5pt]{%
  \begin{tikzpicture}
  \draw[fill] (0,0)-- (90:#1) arc (90:270:#1) -- cycle ;
  \draw[thick] (0,0) circle (#1);
  \end{tikzpicture}}

\DeclareMathOperator*{\argminB}{argmax}

\usepackage{booktabs}
\usepackage{amssymb}
\iccvfinalcopy 

\def\iccvPaperID{22} 

\usepackage{pgfplots}
\pgfplotsset{width=8cm,height = 3.5cm,compat=1.9}

\usepackage[breaklinks=true,bookmarks=false]{hyperref}
\begin{document}

\title{BLACK-BOX ATTACKS ON IMAGE ACTIVITY PREDICTION AND ITS NATURAL LANGUAGE EXPLANATIONS}
\author{Alina Elena Baia\textsuperscript{1}, Valentina Poggioni\textsuperscript{2}, Andrea Cavallaro\textsuperscript{1,3,4}\\
\textsuperscript{1}Idiap Research Institute, Switzerland,
\textsuperscript{2}University of Perugia, Italy, \\
\textsuperscript{3}École Polytechnique Fédérale de Lausanne, Switzerland, \textsuperscript{4}Queen Mary University of London, U.K. \\
{\tt\small \{alina.baia, a.cavallaro\}@idiap.ch}\\
{\tt\small valentina.poggioni@unipg.it}\\
}

\maketitle
\ificcvfinal\thispagestyle{empty}\fi

\begin{abstract} 

Explainable AI (XAI)  methods aim to describe the  decision process of deep neural networks. Early XAI methods produced visual explanations, whereas more recent techniques generate  multimodal  explanations that include textual information and visual representations. Visual XAI methods have been shown to be vulnerable to white-box and gray-box adversarial attacks, with an attacker having full or partial knowledge of and access to the target system. As the vulnerabilities of multimodal XAI models have not been examined, in this paper we assess for the first time the robustness to black-box attacks of the natural language explanations generated by a self-rationalizing image-based activity recognition model. We generate unrestricted, spatially variant perturbations that disrupt the association between the predictions and the corresponding explanations to mislead the model into generating unfaithful explanations. We show that we can create adversarial images that manipulate the explanations of an activity recognition model by having  access only to its final output.
\end{abstract}

\section{Introduction}

Deep neural models are generally black-box systems whose decision-making process is obscure. Explainable artificial intelligence (XAI) aims to make decisions of deep neural models transparent, i.e.~understandable 
by a human~\cite{XAIsurvey}. An XAI model provides  insights into the decision-making process identifying feature importance 
contribution that facilitates error analysis and the identification of uncertain cases.
\blfootnote{\noindent A. Cavallaro acknowledges the support of the CHIST-ERA program through the project GraphNEx, under UK EPSRC grant EP/V062107/1. For the purpose of open access, the author has applied a Creative Commons Attribution (CC BY) license to any Author Accepted Manuscript version arising.
} 
XAI systems favor the assessment of the vulnerabilities  of a model~\cite{XAIpeeking} and interactions with people to support their decisions~\cite{helpmehelptheai}. 
\begin{figure}[t]
  \centering
  \centerline{\includegraphics[width =\columnwidth]{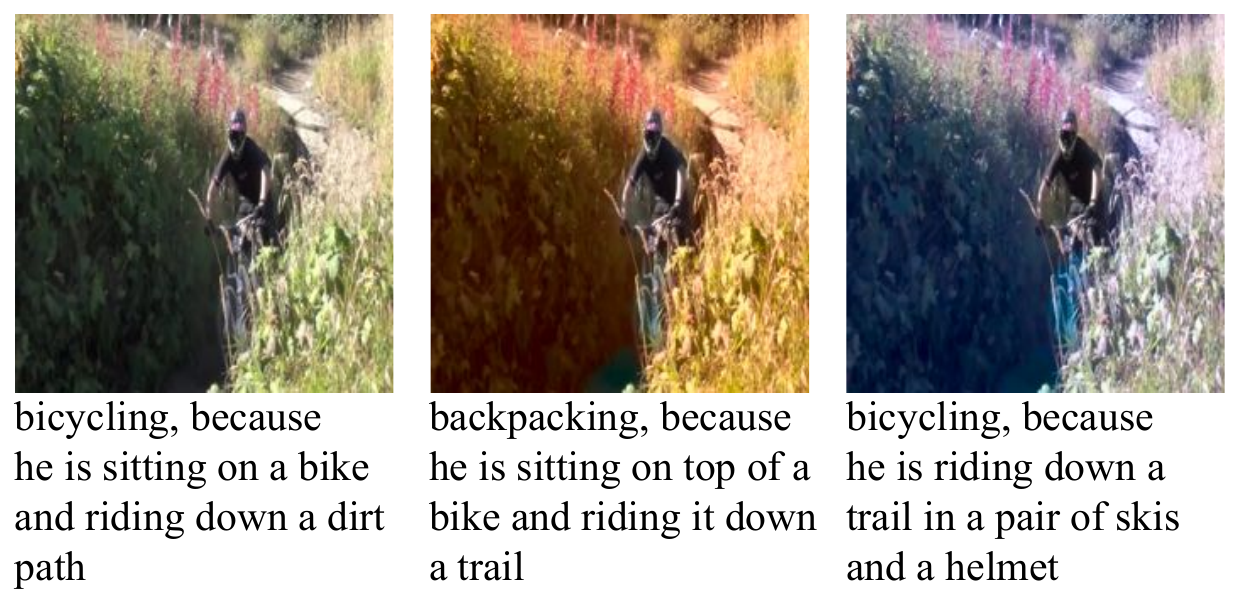}}
\caption{Sample adversarial images generated against NLX-GPT~\cite{sammani2022nlx} 
from a clean image (left) by changing the activity  prediction while maintaining the  textual explanation  (middle) and by maintaining the  activity prediction while changing the  textual explanation (right).}
\label{fig:first_example}
\end{figure}

XAI approaches may generate  visual (V-XAI), textual (T-XAI) or multimodal (M-XAI) explanations. Visual explanations  highlight the most relevant pixel information used by the model~\cite{LIME, selvaraju2017grad, simonyan2013saliency}. Examples include  super-pixels based visualizations (e.g.~LIME~\cite{LIME}), heatmaps~\cite{selvaraju2017grad}, saliency maps~\cite{simonyan2013saliency},  and feature contribution methods inspired by game theory (e.g.~SHAP~\cite{shap}).  However,  V-XAI outputs 
may be difficult to comprehend  for non-expert users when no information is provided on how highlighted pixels influence the decision.
Textual explanations  describe the reasons for a  decision in a more human-interpretable form through natural language~\cite{e-snli-ve,HendricksARDSD16,kayser2021evil,vqaExplaining,marasovic-etal-2020-natural}. Finally,  multimodal explanations  jointly generate textual rationales and visual evidence in the form of attention maps~\cite{park2018multimodaljustifying,sammani2022nlx, wu2018faithfulmultimodal}. A recent self-rationalizing M-XAI method~\cite{sammani2022nlx}  simultaneously predicts the decision and justifies, textually and visually,  what led to that decision. 

Various studies have addressed the vulnerability of V-XAI methods to adversarial attacks~\cite{adebayo2018sanity,DombrowskiAAAMK19, ghorbani2019interpretation, Heo,kindermans2019reliability,whenandhow, slack2020fooling}, however no previous work explicitly considered T-XAI or M-XAI models. 

In this paper we present a black-box\footnote{A black-box attack simulates a realistic threat since there is no need of model-specific information and the access to the target model is limited (i.e.~only its final output).}, content-based and unrestricted\footnote{Unrestricted perturbations allow for more freedom in modifying the image, which improve attack effectiveness and transferability~\cite{Edgefool, Filterfool,shamsabadi2020colorfool, ala,ACE_journal}, and can evade defense mechanisms more effectively~\cite{shamsabadi2020colorfool, DemiguiseAC}.} attack against a natural language XAI model for image classification~\cite{sammani2022nlx}.  We generate the adversarial attack against a vision-language model with unrestricted semantic colorization. Our attack uses only the final output, namely the textual output or/and the visual maps of the model to determine the adversarial perturbations.
We consider two attack scenarios, namely changing the activity prediction while keeping the textual explanation similar and maintaining the same activity prediction while changing the textual explanation (see Figure~\ref{fig:first_example}). 
To  the best of our knowledge, no related work explicitly performs black-box attacks on the prediction-generation mechanism of a natural language-based explanation system. 

In summary, our contributions are as follows:
\begin{itemize}
    \item We propose the first black-box attack against the prediction-explanation mechanism of a natural language explanation model for image classification. We evaluate the robustness of the target model against adversarial image colorization  techniques under two scenarios: changing the prediction while keeping the explanation similar, and keeping the same prediction while changing the explanation.
    
    \item We create adversarial examples  by combining  image semantics  and the information provided by a visual explanation map  to localize the most relevant areas for the prediction and to adapt to different image regions.
    
\end{itemize}


\begin{table}[t!]
\setlength\tabcolsep{5.7pt}
\footnotesize
\centering
\caption{Adversarial attacks against vision-language models. KEY -- \tikzcirclewhite{2.5pt}:~white-box, \tikzcircleblack{2.5pt}:~black-box, \halfcirc{2.5pt}:~gray-box, $\text{T}$:~targeted, $\overline{\text{T}}$:~untargeted, $\text{R}$:~restricted, $\overline{\text{R}}$:~unrestricted, IC:~image captioning, SG:~story ending text generation, IG:~image generation, VQA:~visual question answering, ACT-X:~activity recognition explanation.}
\label{tab:sota_summary}
\begin{tabular}{lcclccccc}
\toprule
 Reference & {Task} & {Box} & $\text{R}$ & $\overline{\text{R}}$ & $\text{T}$  & $\overline{\text{T}}$ \\
\toprule

Chen et al.~\cite{chen2018-show-and-fool} & IC  &   \tikzcirclewhite{2.5pt}  & \checkmark  &  & \checkmark & \\

Zhang et al.~\cite{zhang2020Imagination} & IC &  \tikzcirclewhite{2.5pt}  & \checkmark  & & \checkmark  &  \\

Ji et al.~\cite{ji2020accuracy-preserving}  & IC  & \tikzcirclewhite{2.5pt} & \checkmark &  & \checkmark & \\

Kwon et al.~\cite{kwon2022_restrictedarea} & IC  &  \tikzcirclewhite{2.5pt} & \checkmark &  & & \checkmark  \\
Xu et al.~\cite{xu2019exact} & IC  &  \tikzcirclewhite{2.5pt}  & \checkmark  &  & \checkmark  & \checkmark \\
Bhattad et al.~\cite{Bhattad2020UnrestrictedAE} & IC   & \tikzcirclewhite{2.5pt} & & \checkmark & \checkmark&   \\
Wu et al.~\cite{wu2022AdvGan} & IC &  \tikzcirclewhite{2.5pt},\halfcirc & \checkmark  & &  & \checkmark \\
Wang et al.~\cite{Wang2022UAPGAN} & IC &   \tikzcirclewhite{2.5pt},\halfcirc & \checkmark &  & \checkmark & \checkmark  \\
Sharma et al.~\cite{sharma2018attendandattack} & VQA &  \tikzcirclewhite{2.5pt} & \checkmark &  &  & \checkmark  \\
Huang et al.~\cite{storyendingattack} & SG & \tikzcirclewhite{2.5pt} & \checkmark  &   & &\checkmark \\
Xu et al.~\cite{xu2018fooling}  & IC, VQA & \tikzcirclewhite{2.5pt}  & \checkmark &  & \checkmark &  \\

Lapid et al.~\cite{iseedeadpeople} & IC & \halfcirc & \checkmark  &   & \checkmark &\checkmark \\
Aafaq et al.~\cite{Aafaq} & IC & \halfcirc & \checkmark  &   & \checkmark & \\
Chaturvedi et al.~\cite{chaturvedi2020mimic} & IC, VQA  & \halfcirc & \checkmark  &  & \checkmark & \\

Zhao et al.~\cite{zhao-bb-VLM} & IC, VQA, IG & \tikzcircleblack{2.5pt} & \checkmark  &   & \checkmark & \\

{\bf Ours} & ACT-X & \tikzcircleblack{2.5pt} &  & \checkmark & & \checkmark \\ 

\bottomrule
\end{tabular}
\end{table}

\section{Related works}

V-XAI models are susceptible to adversarial attacks that may, for example, preserve the prediction of the original image but change the explanation~\cite{adebayo2018sanity, DombrowskiAAAMK19, ghorbani2019interpretation, Heo, kindermans2019reliability, slack2020fooling, whenandhow}. Examples of attacks include restricted adversarial perturbations~\cite{ghorbani2019interpretation}, structured manipulations that change the explanation maps to match an arbitrary target map~\cite{DombrowskiAAAMK19},  and adversarial classifiers~\cite{slack2020fooling} that  fool post-hoc explanations  methods such as LIME~\cite{LIME} and SHAP~\cite{shap}. Other works use simple constant shift transformation of the input data~\cite{kindermans2019reliability}, model parameter randomization and data randomization~\cite{adebayo2018sanity}, and  network fine-tuning with adversarial loss~\cite{Heo} to manipulate visual explanations models.

Table~\ref{tab:sota_summary} shows a summary of existing attacks on vision-language models.
Several studies covered V-XAI methods, however no work has yet explicitly considered   textual explanations of  self-rationalizing multimodal explanations models. 
Existing similar research on vision-language models focuses on attacking image captioning or visual question answering models. The attacks use $L_p$-norm restricted perturbations 
and are primarily conducted in a white-box~\cite{chen2018-show-and-fool,ji2020accuracy-preserving, kwon2022_restrictedarea,sharma2018attendandattack,Wang2022UAPGAN, wu2022AdvGan,xu2018fooling, xu2019exact, zhang2020Imagination} or gray-box~\cite{Aafaq, chaturvedi2020mimic, iseedeadpeople, Wang2022UAPGAN, wu2022AdvGan} setup. These attacks are less practical in a real-world scenario since they require prior 
knowledge about the victim model, which is not readily available, and are often designed for specific model architectures. Also, restricted perturbations are often not semantically meaningful~\cite{sparsefool, Su2019OnePA} and can create visible artifacts that can be detected by defenses~\cite{JPEG, sharif2018suitability, Xu2018FeatureSD}. 

%
Attacks on image-to-text generation models may treat the structured output as a single label and design the attack as a targeted complete sentence~\cite{ Aafaq, iseedeadpeople,xu2018fooling}. This idea was extended to targeted keywords attacks that encourage the adversarial caption to include a predefined set of keywords 
in any order~\cite{chen2018-show-and-fool,zhang2020Imagination} or at specified positions in the caption~\cite{xu2019exact}. 
Methods may mask out targeted keywords while preserving the caption quality for the visual content~\cite{ji2020accuracy-preserving}.  Untargeted attacks may use attention maps of the underlying target model to focus the adversarial noise on the regions attended by the model~\cite{sharma2018attendandattack}.  Generative adversarial models have also been used to create adversarial perturbations~\cite{Aafaq,Wang2022UAPGAN,wu2022AdvGan}. Alternatively, adversarial images
may be generated by perturbing an image so that its features  resemble those of a target image forcing the model to output the same caption~\cite{Aafaq,chaturvedi2020mimic,iseedeadpeople}.

Multimodal vision-language models for classification tasks are vulnerable to white-box and gray-box adversarial perturbations on a single modality~\cite{yang2021defending_singlesource} (i.e.~image input) or multiple modalities~\cite{evtimov2020adversarial, readingisntbelieving,zhang2022co-attack} (i.e.~image and text input).
A multimodal white-box iterative attack~\cite{storyendingattack} that fuses image and text modalities attacks has also been introduced to change the output sentence of a multimodal story-ending generation model. A recent black-box attack~\cite{zhao-bb-VLM} deceives large vision-language models assuming a targeted adversarial goal. First, a surrogate model is used to craft adversarial examples with restricted perturbations  and  transfers the adversarial examples to the victim model; then a query-based attacking strategy  generates responses more similar to the targeted text.
 
In this work, we focus on 
a self-rationalizing model and empirically analyze the robustness against black-box content-based unrestricted  attacks by changing either the activity prediction or the explanation. We do not consider the scenario of attacking both activity and explanation since this would be similar to image-captioning attacks that aim to change the entire textual output of a model. The proposed methodology uses only the final decision of the explanation model and does not rely on any surrogate models. Moreover,  considering the attack scenarios, our problem is more challenging since multiple conditions need to be satisfied for an attack to be successful. 

\section{Methodology}
\subsection{Problem definition}

Let I $\in \mathbb{R}^{h \times w \times 3}$ be an RGB image with height $h$ and width $w$. Let  $M_{E}$ be an encoder-decoder M-XAI model such that
$M_E(I) = \bigl( a, e, I_{e} \bigr)$, where  $a= (a_1, a_2, \dots, a_p)$ represents the generated textual description of the activity and $e = (e_1, e_2, \dots, e_n)$ is the generated textual explanation that justifies the activity decision; 
$a_i$ and $e_j$ are words, and  $p$ and $n$ are variable sentence lengths, which depend on the type of activity illustrated in the image $I$. The set of possible activities is not fixed if $M_E$ uses as decoder a language prediction model that  generalizes to activity categories  unseen  during training. 
$I_{e}$ is the visual explanation map generated for the predicted activity using the cross-attention weights of $M_E$.


We define an adversarial example for the explainable model $M_{E}$,  the image $\hat{I}$, such that $M_E(\hat{I}) = \bigl(  \hat{a}, \hat{e}, \hat{I_{e}} \bigr)$, where $\hat a$, $\hat e$, and $\hat{I}_e$ are the activity prediction, textual explanation, and visual explanation generated for the image $\hat I$.  In this work, we focus on the textual explanations and we do not set any conditions on $\hat{I}_e$.
Under the assumption of  faithful explanations (i.e.~explanations that accurately reflect the process behind a prediction)
the label-rationale should be strongly associated~\cite{wiegreffe-etal-2021-measuring}: changing the activity prediction implies a change in its explanation. 
Therefore, our objective is to break the correlation between activity prediction and its explanation by changing one part while keeping the other unchanged.

We therefore consider two attack scenarios, namely $S1$ for which the activities predictions are different ($a \neq \hat{a}$), but  the explanations are similar ($e \simeq \hat{e}$), and $S2$, for which the activities predictions are the same ($a = \hat{a}$), but the explanations are different ($e \nsim \hat{e}$).



\subsection{Black-box unrestricted attacks}
\label{subsec:bb}

We condition the perturbation generation on the activity prediction and textual explanation. We craft region-specific unrestricted perturbations and generate adversarial examples following the image semantics-based idea proposed in~\cite{shamsabadi2020colorfool}. To determine the adversarial perturbations we use the (dis)similarity between textual explanations.  We consider two strategies for perturbing the semantic areas accordingly and extend them to our problem. The first strategy is a 
random colorization approach~\cite{shamsabadi2020colorfool} and the second is a strategy  that combine photo editing techniques~\cite{Baia2021EffectiveUU}.

\noindent {\bf Explanation similarity.}  We measure the difference between $e$, the textual explanation generated for the clean image $I$, and $\hat{e}$, the explanation generated for the perturbed image $\hat{I}$. Let $E(\cdot)$ be a transformer-based network~\cite{sentenceTransformer} that computes the vector embedding of a sentence. Then we calculate  
the similarity  between $e$ and $\hat{e}$, $Q_{\hat{T}}(I, \hat{I})$,  as the cosine similarity\footnote{We use a cosine-similarity  measure with neural sentence embedding because of its highest correlation with human judgement~\cite{clinciu-etal-2021-study, kayser2021evil, sentenceTransformer} and out-performance of other methods such as METEOR~\cite{METEOR} or BLEU~\cite{BLEU}.}  normalized in the range [0,1]:
\begin{equation}
    Q_{\hat{T}}(I, \hat{I})  =\frac{1}{2} \left (
 \frac{\sum_{i=1}^{n}{E(e)_i E(\hat e)_i}}{\sqrt{\sum_{i=1}^{n}{{E(e)^2_i}}}\sqrt{\sum_{i=1}^{n}{{E(\hat e)^2_i}}}} +1 \right),
 \label{eqn:cosine}
\end{equation}
where $n$ is the size of the embedding vector. The larger the similarity $Q_{\hat{T}}(I, \hat{I})$, the more similar the explanations for $I$ and $\hat{I}$. For example, let us consider the sentences $e_1$: \textit{he is standing on a bridge with a backpack on his back}, $e_2$: \textit{he is wearing a backpack and standing on a bridge}, and $e_3$: \textit{he is standing in a field with a frisbee in his hand}. Sentences $e_1$ and $e_2$ have
the same meaning and their  similarity is 0.97. Sentences $e_1$ and $e_3$ describe different scenarios (although they share a few words) and their difference is reflected in a lower  similarity of 0.69. Sentences $e_2$ and $e_3$  also have a low similarity of 0.70.

\begin{figure}[t]
  \centering
    \begin{subfigure}{0.32\columnwidth}
    \centering
      \includegraphics[width=\columnwidth]{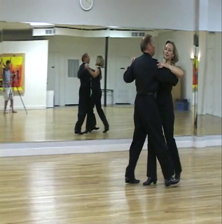}
    \end{subfigure}
    \begin{subfigure}{0.32\columnwidth}
        \centering
      \includegraphics[width=\columnwidth]{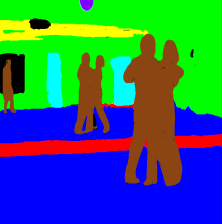}
    \end{subfigure}
    \begin{subfigure}{0.32\columnwidth}
        \centering
      \includegraphics[width=\columnwidth]{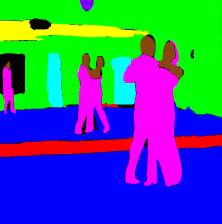}
    \end{subfigure}    
  \caption{Example of semantic regions obtained after the first step (middle) and last step (right) of the multi-step segmentation scheme. Regions in brown are considered sensitive to color changes. }
\label{fig:example_s2}
\end{figure}

 
\noindent {\bf Image partitioning.}  We use a multi-step segmentation approach to partition an image 
into sensitive regions, $R_{i}^s$, and  non-sensitive regions, $R_{j}^n$. Sensitive regions correspond to objects whose unrealistic colors and appearance could raise suspicion (e.g.~human skin), whereas non-sensitive regions can have their colors arbitrarily modified  without necessarily making the image look unnatural.
We represent an image $I$ as:
\begin{equation}
I = \bigcup R_{i}^s \cup \bigcup R_{j}^n.     
\end{equation}
 First, we use semantic segmentation to partition an image into semantic regions, such as person, sky, car, building~\cite{mask2former}. Next, we detect skin\footnote{$\text{Skin Segmentation Network}$:\url{https://github.com/WillBrennan/SemanticSegmentation}}
 areas on top of semantic regions representing people and mark the skin as sensitive and unalterable. 
 Finally, we further partition each semantic region into smaller areas and obtain the non-sensitive regions with color-based oversegmentation~\cite{automaticfuzzyclustering}.  An example is shown in Figure~\ref{fig:example_s2}.



\noindent {\bf Optimization process.}  
We find an adversarial example for ${I}$ in  $S1$, $\hat{I}_{S1}$, whose generated explanation has the highest similarity with that generated for $I$, while also having a different activity prediction, as follows:
\begin{equation}
\label{eqn:optimization_s1}
    \hat{I}_{S1} = \argminB_{\hat{I}}\bigl(Q_{\hat{T}}(I, \hat{I})   \mathbbm{1}_{\{(a, \hat{a}): a \neq \hat{a}\}},  Q_{\hat{I}}(I, \hat{I})\bigr),
\end{equation}
where $Q_{\hat{I}}(I, {\hat{I}})$ is used to reduce the noticeability of the perturbation and is implemented as SSIM~\cite{SSIM} between the clean image, $I$, and the candidate adversarial example, ${\hat{I}}$, and $\mathbbm{1}_{\{(a, \hat{a}): a \neq \hat{a}\}}$ is the indicator function whose value is 1 only if the predicted 
activity of $\hat{I}$ is different from the activity of $I$.

Similarly, we find an adversarial example for ${I}$ in  $S2$, $\hat{I}_{S2}$, whose generated explanation has the lowest  similarity with the explanation generated for the clean image $I$, while also having the same the activity prediction as $I$, as:
\begin{equation}
\label{eqn:optimization_s2}
    \hat{I}_{S2} = \argminB_{\hat{I}}\bigl(1 - Q_{\hat{T}}(I, \hat{I})  \mathbbm{1}_{\{ (a, \hat{a}): a = \hat{a}\}}, Q_{\hat{I}}(I, \hat{I})\bigr),
\end{equation}
where $\mathbbm{1}_{\{ (a, \hat{a}): a = \hat{a}\}}$ is the indicator function whose value is 1 only if the predicted 
activity of $\hat{I}$ is the same as the activity of $I$.

\noindent {\bf Random colorization.} 
We extend ColorFool~\cite{shamsabadi2020colorfool} to consider the explanation similarity $Q_{\hat{T}}$, as defined in Eq.~\ref{eqn:cosine}. 
We refer to this method as ColorFoolX (CFX).
In this case, we do not use the image similarity $Q_{\hat{I}}$ in the process of finding the adversarial example. We rely only on the image region semantics and prior information about color perception in each region to generate the adversarial images. 
ColorFool  uses the semantic regions computed in the first step of the multi-step segmentation scheme and defines four types of sensitive regions: person, sky, vegetation, and water.
Adversarial images are generated by randomly modifying the {\em a} and {\em b} components of the regions in the perceptually uniform {\em Lab} color space within specific color ranges, which depends on the semantics of a region, without changing the lightness {\em L}. ColorFool avoids perturbing regions representing people. 


\noindent {\bf Combining  editing filters.} 
We extend a combination of image editing filters method~\cite{Baia2021EffectiveUU} to perform localized attention-based attacks. The method manipulates image attributes like saturation, contrast, brightness, sharpness, and applies edge enhancement, gamma correction or soft light gradients. We restrict the perturbations to non-sensitive areas $R_{j}^n$  using the information from $I_e$: we select the non-sensitive areas that are the most important for the activity prediction, $R_{a}^n$, for  $S1$, and the least important non-sensitive areas for the activity prediction, $R_{na}^n$, for  $S2$. 

We generate $\hat{I}$, through a sequence of $L$ filters on $I$, for $S1$ as:
 \begin{equation}
    \hat{I}=  R_{i}^s \cup f^{\alpha_{t_1},\beta_{t_1} }_{t_1} \circ f^{a_{t_2}, \beta_{t_2}}_{t_2} \circ \dots \circ f^{a_{t_L}, \beta_{t_L}}_{t_L}(R_{a}^n) \cup R_{na}^n,
\end{equation}
and for $S2$ as:
 \begin{equation}
    \hat{I} =  R_{i}^s \cup R_{a}^n \cup  f^{\alpha_{t_1},\beta_{t_1} }_{t_1} \circ f^{a_{t_2}, \beta_{t_2}}_{t_2} \circ \dots \circ f^{a_{t_L}, \beta_{t_L}}_{t_L}(R_{na}^n),
\end{equation}
where each $f_{i}^{\alpha_i, \beta_i}$ is selected from a set of $F$ predefined filters parameterized with  $\beta_i$ that controls the 
amount of change of each property (intensity), and $\alpha_i$, the parameter of the alpha blending between the clean image and the filtered image. The optimal filter configuration is found with a nested evolutionary algorithm consisting of an outer optimization step that determines the sequence of $L$ filters with $f_{t_{i}} \in F$ with a genetic algorithm (GA)~\cite{GA}, and an inner optimization step that determines  the values of $(\alpha_{t_{i}}, \beta_{t_{i}})$
of each selected filter in the outer step with an Evolutionary Strategies (ES)~\cite{ES}. 

We consider both $Q_{\hat{T}}$ and $Q_{\hat{I}}$ functions to find the adversarial examples, as defined in Eq.~\ref{eqn:optimization_s1},~\ref{eqn:optimization_s2}. Considering the conflicting nature of the two functions we formulate the optimization process as multi-objective optimization, handled by the NSGA-II algorithm~\cite{NSGAII}, to find the best trade-off between $Q_{\hat{T}}$ and $Q_{\hat{I}}$.

\section{Validation}
\label{sec:validation}
\subsection{Experimental setup}
\noindent {\bf Multimodal explanation model.}  We perform the attacks on the multimodal explanation model NLX-GPT for activity recognition~\cite{sammani2022nlx}, which textually explains its prediction using CLIP~\cite{CLIP} as vision encoder and the distilled GPT-2 pre-trained model~\cite{sanh2019distilbert, brown2020language}
 as a decoder.  NLX-GPT generates also a visual explanation map based on the cross-attention weights of the model. The distilled GPT-2 was pre-trained on image-caption pairs (COCO captions~\cite{COCOcaptions}, Flickr30k~\cite{flickr30k}, visual genome~\cite{visualgenome} and image-paragraph captioning~\cite{paragraphcaptions}). 
NLX-GPT  was fine-tuned on the activity recognition dataset ACT-X~\cite{park2018multimodaljustifying} (18k images). The 
encoder is fixed for both the pre-training and fine-tuning stages.

\noindent {\bf Dataset.}  We use the test set of the ACT-X~\cite{park2018multimodaljustifying}, a 3,620-image dataset used to explain  decisions of activity recognition models. Each image is labeled with an activity and three explanations. We perform the attack on the 1,829 images with correctly predicted activity by NLX-GPT.  

\noindent {\bf Cases.} We compare different filtering  approaches and objective functions. We analyze the following cases: full image filtering (FL-s) and localized filtering (LC-s,  as described in Section~\ref{subsec:bb}) with single objective ($Q_{\hat T}$) for explanation (dis)similarity; full image filtering (FL-m) and localized filtering (LC-m) with multi-objective function ($Q_{\hat T}$, $Q_{\hat I}$), and ColorFoolX (CFX). Note that CFX does not account for image similarity during the attack.

\noindent {\bf Parameters.} 
For CFX we allow a maximum of 1000 trials. For FL-s and LC-s we follow the CFX iterative approach. For FL-m and LC-m we use the multi-objective evolutionary optimization with the configuration proposed in~\cite{Baia2021EffectiveUU}. We set the size of the outer population to $N_{out} = 10$, the number of outer generations to $G_{out} =10$, and the mutation probability to $\rho = 0.5$. The inner population size is $\lambda = 5$, inner generations $G_{in} = 3$ with initial learning rate $lr = 0.1$ and decay rate $\beta = 0.75$.

\subsection{Performance evaluation}

\noindent {\bf Success of the attacks.}
We measure the success rate, $S_r$, of the adversarial attacks as:
\begin{equation}
\label{eqn:sr}
    S_{r}= \frac{1}{N_a}\sum\nolimits_{j=1}^{N_{a}} \mathbbm{1}_{\omega},
\end{equation}
where $N_a$ is the total number of images and, for  $S1$: 
\begin{equation}
\label{eqn:above_thrs}
    \omega  \triangleq  \{ (a_j, \hat a_j): a_j \neq \hat a_j  \land Q_{\hat{T}}(I_j, \hat{I}_j) \geq t \},
\end{equation}
where $t$ is a threshold; and, for  $S2$: 
%
\begin{equation}
\label{eqn:under_thrs}
\omega  \triangleq  \{ (a_j, \hat a_j): a_j = \hat a_j  \land Q_{\hat{T}}(I_j, \hat{I}_j) < t \}. 
\end{equation}

\begin{figure}[t]
\begin{tikzpicture}
\begin{axis}[
    title={},
    xlabel={Explanation groups},
    ylabel={Similarity class},
    xmin=0, xmax=10,
    ymin=0, ymax=6,
    xtick={1, 2, 3, 4, 5, 6, 7, 8, 9},
    xticklabels={G1, G2, G3, G4, G5, G6, G7, G8, G9},
    ytick={1,2,3,4,5},
    yticklabels={C1,C2,C3,C4,C5},
    legend pos=north west,
    ymajorgrids=true,
    grid style=dashed,
]

\addplot[
    only marks,
    color=blue,   
    mark=square,
    mark size=2.9pt]
    coordinates {
    (1,5) (2,4) (3,4) (4,3) (5,2) (6,2) (7,1) (8,1) (9,1)
    };
    \legend{}
    
\end{axis}
\end{tikzpicture}
\caption{Mapping between explanation groups  and similarity classes. KEY -- C1: not similar at all, C2: a little similar, C3: somehow similar, C4: very similar, C5: they are the same. 
Explanations pairs with $Q_{\hat{T}} >0.85$ (i.e. G1-G3) are rated as highly similar.}
\label{fig:groups_vs_similarity}
\end{figure}
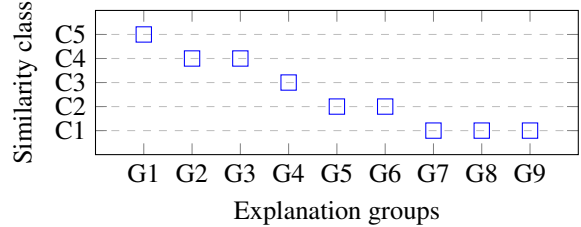

%
We determined the value of $t$ with a subjective human evaluation of the similarity  of explanations pairs. We created nine groups for the explanations based on their  similarity, such that $G_{i} = \{(e, \hat e): Q_{\hat{T}} \in ( 1-0.05i, 1-0.05(i-1)]  \}$ with $i \in \{1,2,\dots, 8 \}$ and $G_{9} = \{ (e, \hat e): Q_{\hat{T}} \in (0, 0.6]\}$. From each group, we randomly selected ten $(e, \hat e)$ pairs that were rated on semantic similarity on a 5-level Likert scale: {\em not similar at all}; {\em a little similar}; {\em somehow similar}; {\em very similar}; and {\em they are the same}. We used majority voting  to assign each pair of explanations to a similarity class. Likewise, we labeled each group with the most frequent similarity class of the questions within the group. Eleven people who did not see the data prior to the test rated the similarity and could change their rating before completing the test. The mapping between explanation groups and similarity classes is shown in Figure~\ref{fig:groups_vs_similarity}. We choose \textit{somehow similar} class as similarity breaking point. This similarity class maps to group G4, which corresponds to $Q_{\hat{T}} < 0.85$. Thus, we set the threshold $t = 0.85$. 
%

\noindent {\bf Image quality.} 
We evaluate the quality of the adversarial images with MANIQA~\cite{MANIQA}, a transformer-based no reference image quality assessment metric that won the NTIRE2022 NR-IQA challenge~\cite{NTIRE2022}. MANIQA scores $\in$ [0,1] and the higher the score, the better the quality. 

\noindent {\bf Image colorfulness.} We also analyze the colorfulness~\cite{colorfulness} of the adversarial images 
and compare it with the colorfulness of original images in order to evaluate whether the colorization attacks generate images with 
color  vividness in accordance with human perception. 
Given an RGB image, first the opponent color space representation is computed as:
\begin{equation}
    rg =  R - G, \quad  yb =  \frac{1}{2}(R+G)-B,
\end{equation}
where $R, G, B$ are the red, green, and blue channels. Next, the standard deviation $\sigma$ and the mean pixel values $\mu$ are calculated as:
\begin{equation}
    \sigma =  \sqrt{\sigma^2_{rg} + \sigma^2_{yb}}, \quad \mu =  \sqrt{\mu^2_{rg} + \mu^2_{yb}} .
\end{equation}
Finally, the colorfulness metric is defined as:
\begin{equation}
     C = \sigma+ 0.3 \mu .
\end{equation}
The higher the  $C$ score, the more colorful the image. 
%

\begin{figure}[t]
  \centering

     \begin{subfigure}{\columnwidth}
    \centering
      \includegraphics[width=\columnwidth]{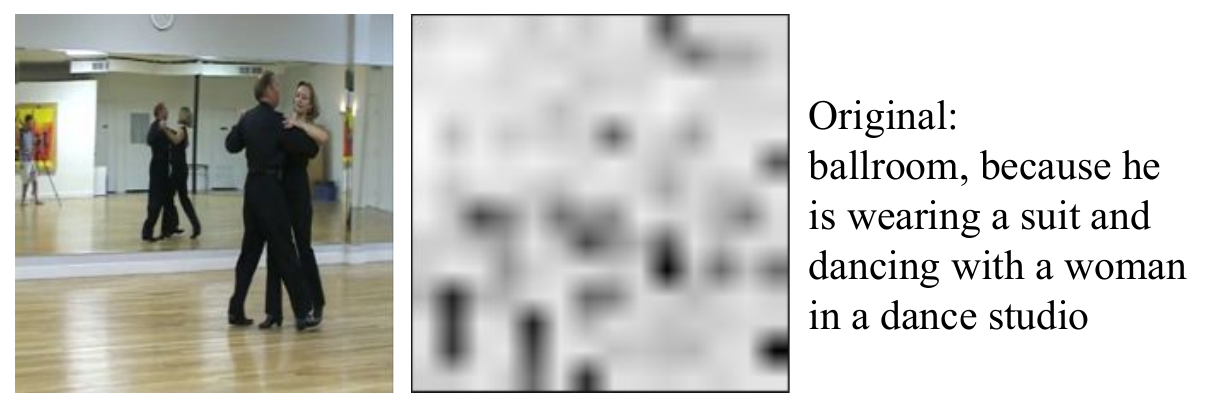}
    \end{subfigure}
    
    \begin{subfigure}{\columnwidth}
    \centering
      \includegraphics[width=\columnwidth]{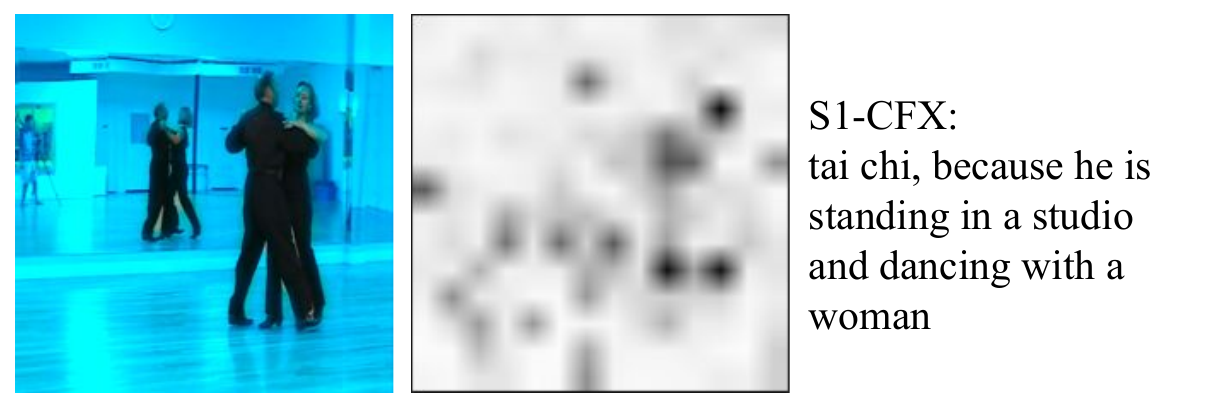}
    \end{subfigure}
        
    \begin{subfigure}{\columnwidth}
    \centering
      \includegraphics[width=\columnwidth]{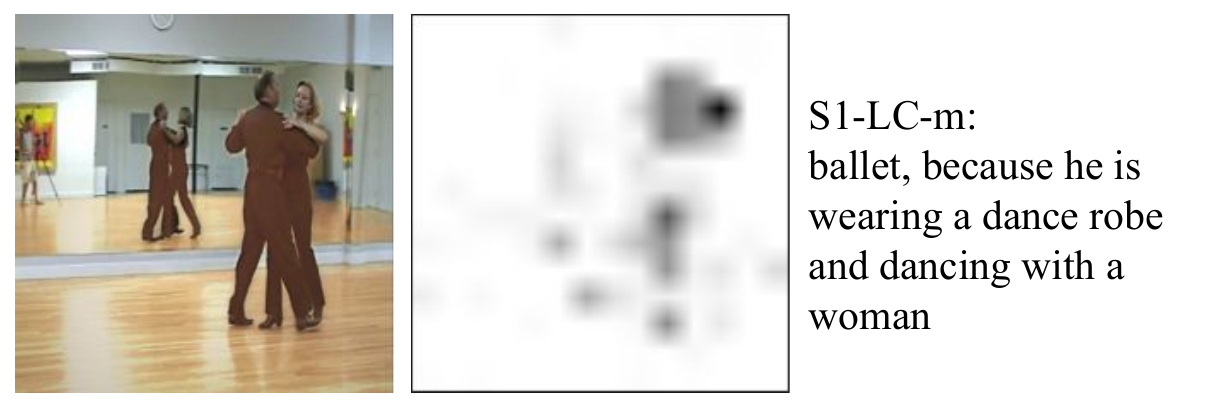}
    \end{subfigure}

    \begin{subfigure}{\columnwidth}
    \centering
      \includegraphics[width=\columnwidth]{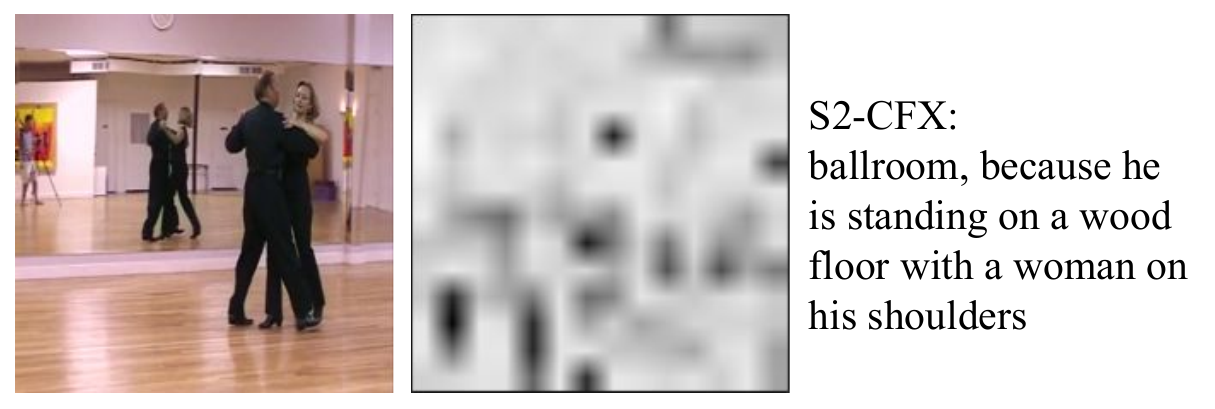}
    \end{subfigure}

    \begin{subfigure}{\columnwidth}
    \centering
      \includegraphics[width=\columnwidth]{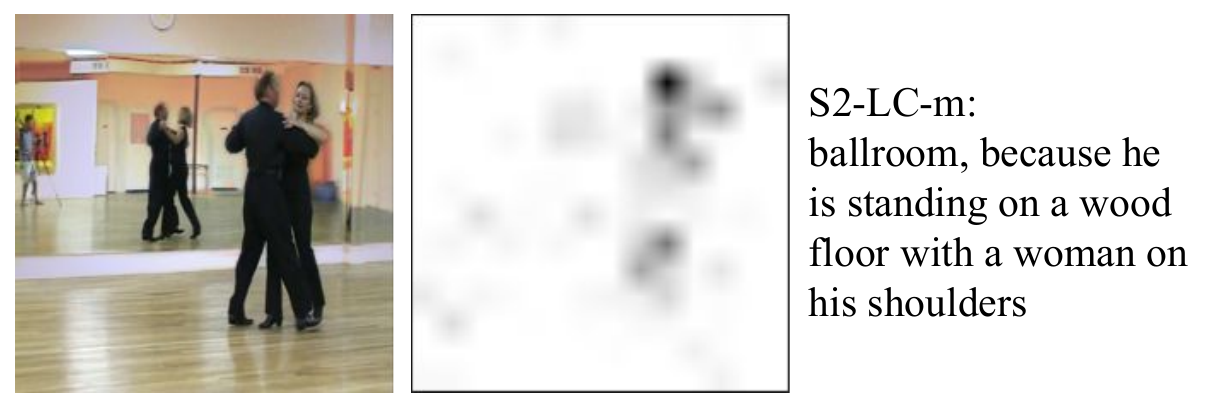}
    \end{subfigure}

  \caption{Adversarial images  generated for a clean image (top left). 
  The visual explanation maps for the activity prediction are shown next to each image. For  $S1$ the images have a different activity and the textual explanations are similar. For $S2$ the images have the same activity but different textual explanations.
  The MANIQA scores for the images are 0.69,   0.63, 0.70, 0.72, 0.64, from top to bottom, respectively. 
  }

\label{fig:example_all}
\end{figure}
\subsection{Results and Discussion}
\noindent {\bf Success of the attacks.} 
Table~\ref{tab:sr_with_treshold} reports the success rates for all methods under both scenarios.
Methods considering only the explanation similarity (i.e. CFX, FL-s) achieve the best success rate with $S_r$ of 64.62\% for CFX and  63.09\% for FL-s for  $S1$
, and $S_r$ of 73.82 \% for CFX and 77.53\% for FL-s in  $S2$. 
CFX and FL-s  apply the perturbation across wider areas of the image than  LC-s, which perturbs small regions selected by combining over-segmentation and visual map information. Moreover, the perturbation is only limited by the semantic region information, which allows more intense modifications than in the case of the multi-objective setup where we use an image similarity metric, $Q_{\hat{I}}$, to calibrate the perturbation. The $S_r$ decreases  as we focus  on more localized areas (LC-s) and as we limit the freedom of the attack with the image similarity function (LC-m). This behavior could  be caused by the noisiness and inaccuracy of the cross-attention visual maps, which may fail to accurately explain visually why the model made a certain decision. Since we use the visual maps to localize the areas to attack,
inaccurate visual maps lead to selecting areas that are irrelevant for the prediction. These model-intrinsic visual attention maps require more investigation to fully assess their relevance for the localized attacks. 
\begin{table}[t!]
\centering
\caption{Success rate (\%) for the two scenarios. 
KEY -- CFX: ColorFoolX, LC-s: localized filtering with a single objective, FL-s: full image filtering with a single objective, LC-m: localized filtering with multi-objective, FL-m: full image filtering with multi-objective.  }
\label{tab:sr_with_treshold}
\begin{tabular}{cccccc}
\toprule
{Scenario} &  CFX & LC-s& FL-s & LC-m & FL-m \\
\midrule
$S1$ & 64.62 & 51.33 & 63.09 & 43.47 & 47.62\\
{$S2$} & 73.82 & 67.47& 77.53 & 51.76 & 49.45\\
\bottomrule
\end{tabular}
\end{table}
We further notice a decrease in attack performance as we enforce an additional constraint on the optimization. On top of the area restriction we also control the applied perturbation using $Q_{\hat I}$. Thus, the algorithm has to find a trade-off 
between explanation (dis)similarity and image similarity. 
The found solution  may sometimes prioritize image similarity over explanation similarity leading to a decrease of the attack success rate.
We also notice that the methods are more effective in $S2$  
achieving a $S_r$ of up to 77.53\% for FL-s. In this scenario, the selected alterable areas are more numerous since we focus on regions that are not highly attended by the explanation model, and thus in general the adversarial perturbation is applied on larger image areas than in the case of LC methods.
Moreover, we observe that in the case of localized attacks, LC-s and LC-m,  the visual attention maps relative to the activity prediction are less noisy and the attention is  primarily focused in one area of the image, whereas for CFX more image regions are attended to, similarly to the original image. Localized attacks, for their nature, are more effective in altering the attention of the model.
\begin{figure*}[ht]
  \centering
    \begin{subfigure}{0.33\textwidth}
    \centering
      \includegraphics[width=\textwidth]{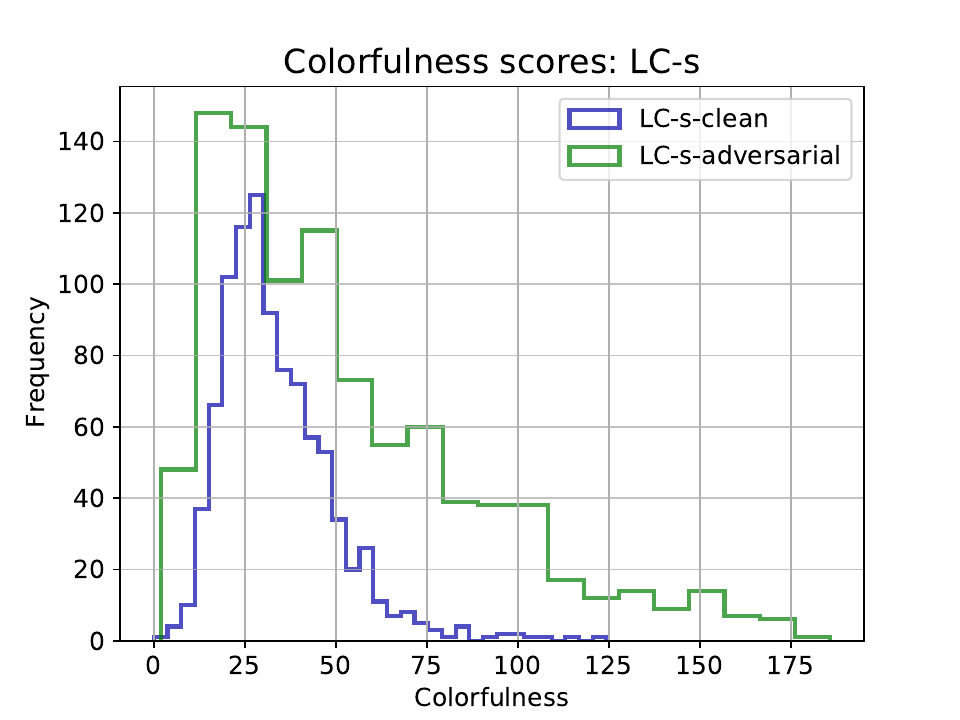}
    \end{subfigure}
    \begin{subfigure}{0.33\textwidth}
        \centering
      \includegraphics[width=\textwidth]{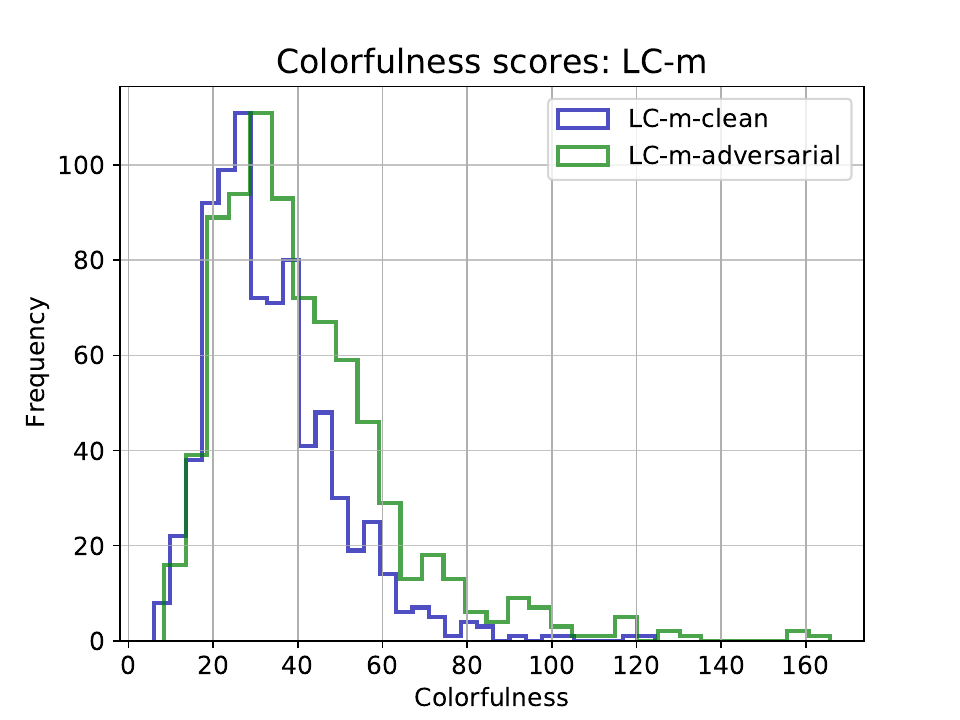}
    \end{subfigure}
    \begin{subfigure}{0.33\textwidth}
        \centering
      \includegraphics[width=\textwidth]{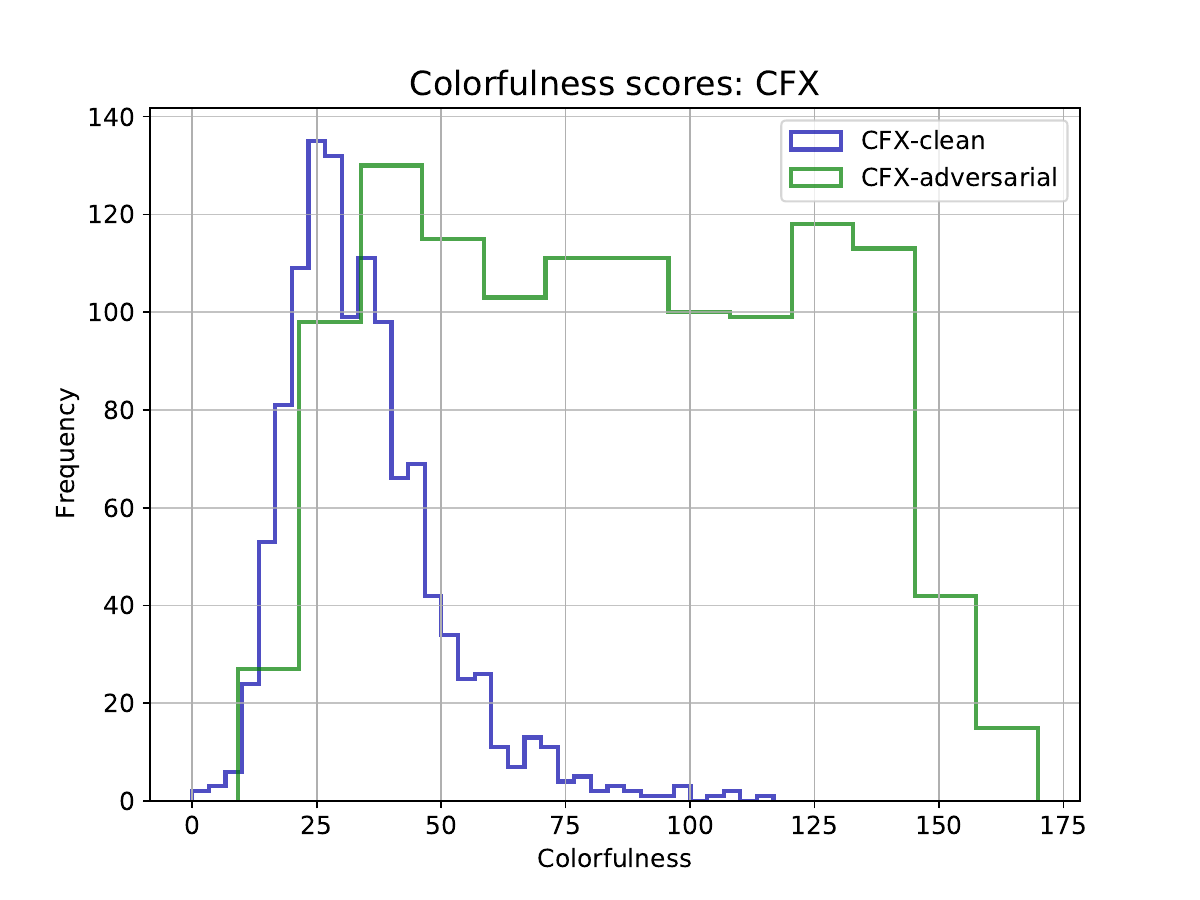}
    \end{subfigure}

        \begin{subfigure}{0.33\textwidth}
    \centering
      \includegraphics[width=\textwidth]{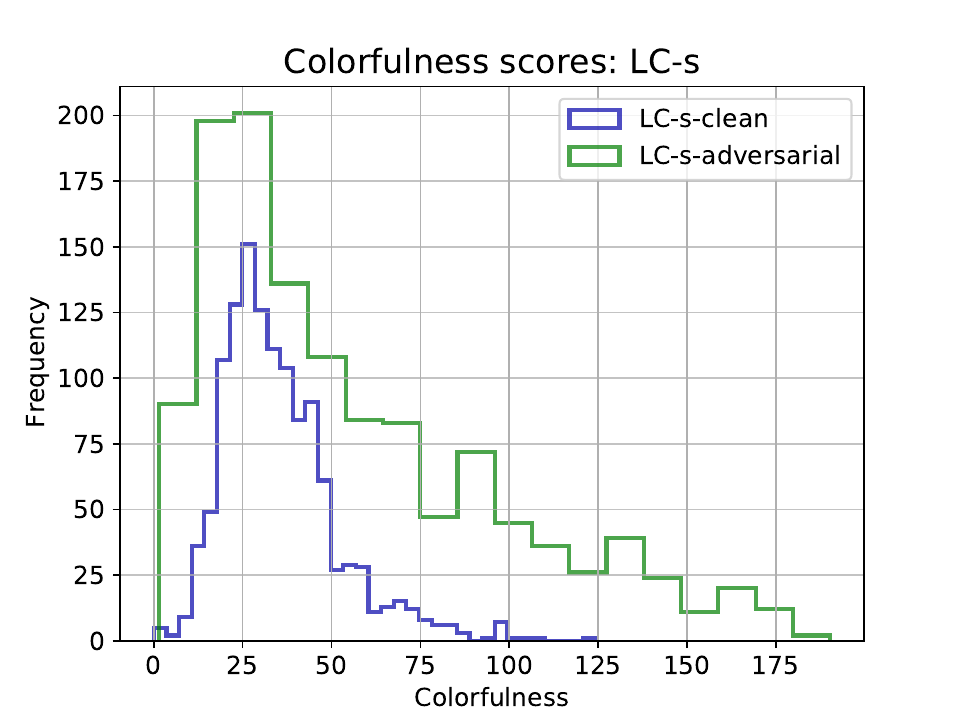}
    \end{subfigure}
    \begin{subfigure}{0.33\textwidth}
        \centering
      \includegraphics[width=\textwidth]{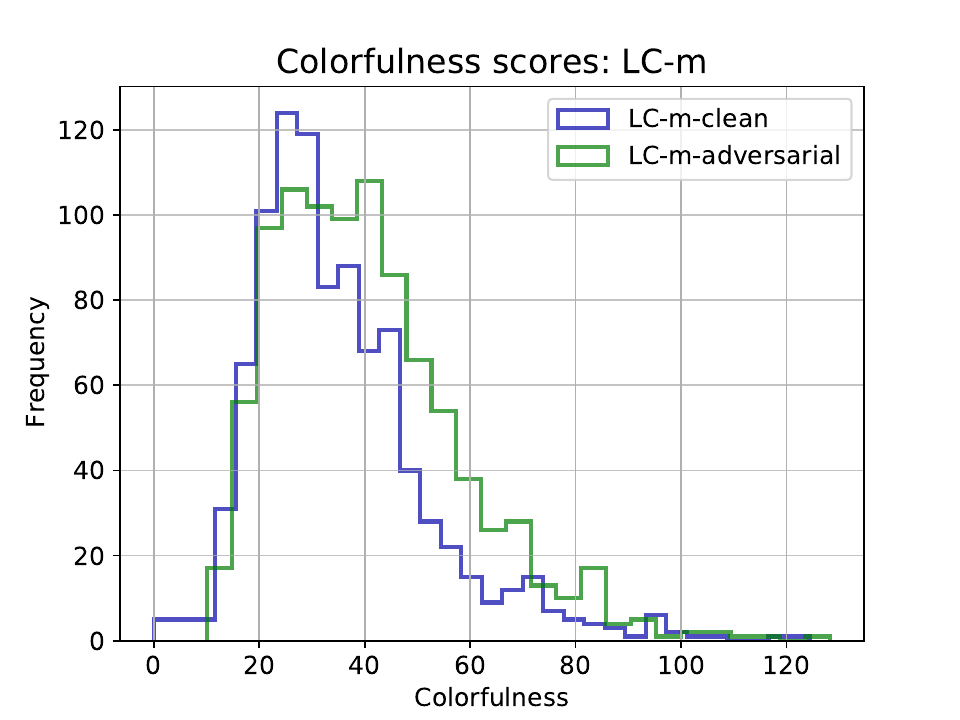}
    \end{subfigure}
    \begin{subfigure}{0.33\textwidth}
        \centering
      \includegraphics[width=\textwidth]{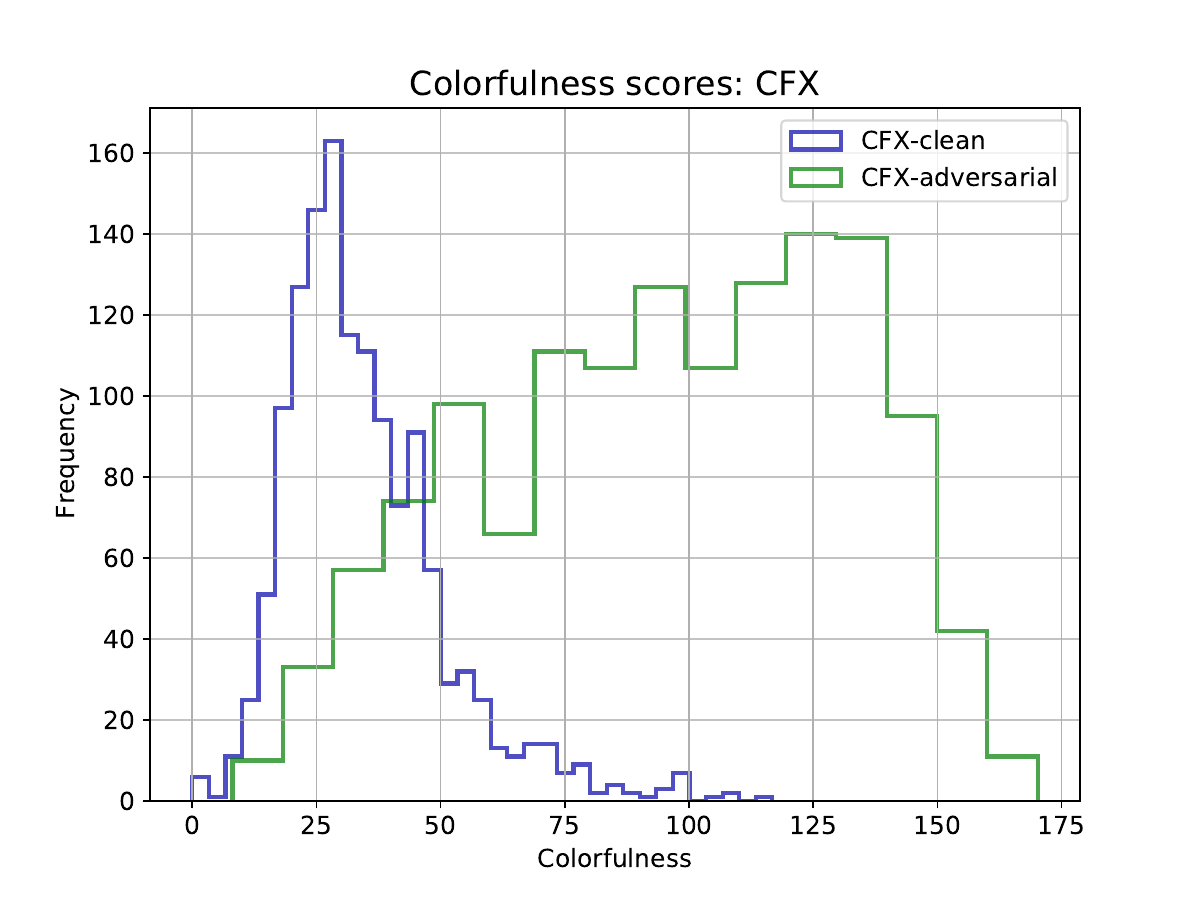}
    \end{subfigure}  
  \caption{Colorfulness scores distribution for  $S1$ (top row) 
  and for  $S2$ 
  (bottom row). The adversarial
examples generated with LC-m and FL-m have colors similar to the original images. In the case of CFX, the colors of adversarial examples diverge from the  distribution of original images. The higher the score, the more colorful the image.}
\label{fig:colorfulness_distribution}
\end{figure*}

\noindent {\bf Image quality.} Both methods produce comparable results, however, the generated adversarial images have different visual characteristics and aesthetics. In general, the image filtering attacks produce images with more toned-down soft vintage looks while most of the images generated by CFX have vivid colors (see Figure~\ref{fig:example_all}). Table~\ref{tab:maniqa} reports the average MANIQA and standard deviation scores for the adversarial images and their corresponding clean versions. The average MANIQA score varies from 0.65 for FL-s and LC-s to 0.68 for CFX, LC-m, and FL-m. As a reference, the average score on the clean images is 0.70. This suggests that the adversarial perturbations do not substantially degrade the image quality. 
\begin{table}[t]
\setlength\tabcolsep{3.8pt}
\footnotesize
\centering
\caption{Average (and standard deviation) of MANIQA scores for the adversarial images and their corresponding clean images.}
\label{tab:maniqa}
\begin{tabular}{ccccc}
\toprule
\multirow{2}{*}{\diagbox{Attack}{Scenario}} &\multicolumn{2}{c}{ $S1$} & \multicolumn{2}{c}{$S2$} \\
\cmidrule(lr){2-3}
\cmidrule(lr){4-5}
   &  \multicolumn{1}{l}{Clean} & Adversarial                                               & \multicolumn{1}{l}{Clean} & Adversarial      \\
\toprule
CFX &.70   $\pm$ .05      & .68     $\pm$ .06      &     .70  $\pm$ .05  & .67   $\pm$ .06      \\ 
LC-s &       .69    $\pm$ .05  &     .66  $\pm$ .06    &    .70  $\pm$ .04  & .65  $\pm$ .07       \\ 
FL-s &        .70   $\pm$ .05  &      .65  $\pm$ .07    &   .70  $\pm$ .05  &    .65   $\pm$ .06   \\ 
LC-m & .69        $\pm$ .05      & .67        $\pm$ .07    &     .70   $\pm$ .04& .68    $\pm$ .05     \\ 
FL-m &   .70      $\pm$ .05    & .66         $\pm$ .07   &     .70   $\pm$ .05 & .68      $\pm$ .05   \\ 

\bottomrule
\end{tabular}
\end{table}

\begin{figure}[t]
\vspace{-0.15cm}
\centering
\includegraphics[width=0.8\columnwidth]{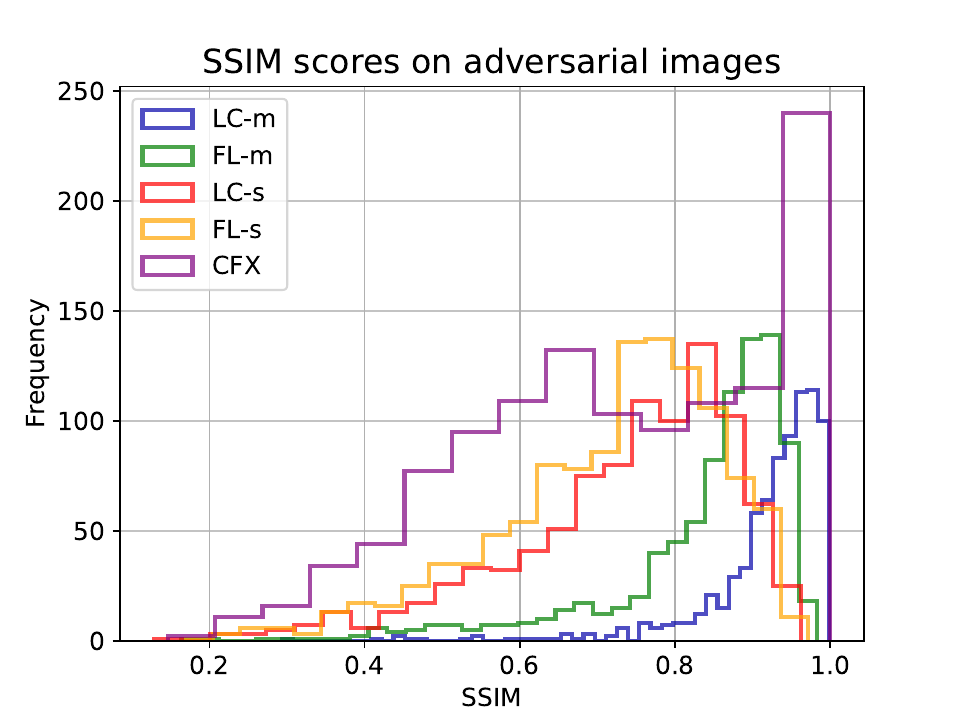}
\caption{SSIM distribution for  $S1$. FL and LC benefit from using the image similarity objective for the optimization. CFX generates images with the highest SSIM values because it does not alter the lightness attribute of the images, which can affect the structural similarity.}
\label{fig:SSIM_scenario1}
\end{figure}

\noindent {\bf Image colorfulness.}  Figure~\ref{fig:colorfulness_distribution}  shows the distribution of colorfulness scores of adversarial images and their corresponding original version. 
LC-m and FL-m generate images with colors most similar to the original images, 
whereas LC-s and FL-s tend to generate images with more faded colors. This indicates that the image similarity objective contributes toward the generations of more natural-looking images, as also shown by the SSIM scores in Figure~\ref{fig:SSIM_scenario1}. On the contrary,
CFX generates very colorful images that  
diverge the most from the original distribution (Figure~\ref{fig:colorfulness_distribution}). 
However, images different from the original ones do not necessarily imply worse quality. Thus, a human subjective evaluation remains the best way to  assess the perceptual realism, which we will address in future work.


\noindent {\bf Ablation study.} 
We perform an ablation study to verify the contribution of each part of the multi-objective function of FL-m and LC-m in both attack success rate and SSIM values (Table~\ref{tab:ablation_study}). 
We start with a random approach, where we randomly perturb the images while only considering changing the activity prediction, disregarding explanation and image similarity.
Then we consider each objective  separately. For the image similarity objective, $Q_{\hat I}$, the aim is to find the image  that changes the activity prediction and has the highest SSIM.
For the explanation objective, $Q_{\hat T}$, the goal is to find an image that changes the activity prediction and has the highest explanation similarity.  
When using both objectives, the goal is to find an adversarial image that has a different activity prediction, high explanation similarity and high image similarity.  
We consider both full image filtering, FL, and localized image filtering, LC. In the case of the image similarity objective only, adversarial images have the highest SSIM scores but a low  $S_r$. However, the textual explanation objective achieves the highest $S_r$ at the expense of the image similarity. This is the main justification for using the version with both objectives to find a trade-off between $S_r$ and SSIM.
\begin{table}[t]
\setlength\tabcolsep{5.5pt}
\footnotesize
    \centering
    \caption{Success rate ($S_r$) and structural similarity (SSIM) for different objective functions for  $S1$.
    }
    \label{tab:ablation_study}
    \begin{tabular}{cccccc}
    \hline
    Text similarity & Image similarity & \multicolumn{2}{c}{LC} &  \multicolumn{2}{c}{FL}\\
    \cmidrule(lr){3-4}
    \cmidrule(lr){5-6}
        &  & $S_r \%$ & SSIM & $S_r \%$ & SSIM\\

    \hline
    & & 21.87&  .86  & 22.86 & .73  \\
    \checkmark & & 51.33 & .73  & 63.09 & .72\\
    &\checkmark &26.93 & .95  & 28.16 & .84 \\
    \checkmark & \checkmark & 43.47 & .92 & 47.62&  .84 \\
    \hline
    \end{tabular}

    \label{avg_SSIM}
    \end{table}

Figure~\ref{fig:SSIM_scenario1} shows the generally large SSIM values of the adversarial examples obtained with  LC-s, FL-s, LC-m, FL-m, and CFX for $S1$.  The results show that using SSIM for the optimization  of the FL and LC is useful for generating images with  higher similarity since the type of modification applied can alter the  structural similarity.
Among all, CFX generates images with the highest SSIM values because it does not directly target the lightness attribute in the images, which can degrade the structural similarity.

We also conducted the analysis for CFX to assess its 
attack capabilities with respect to the original version of ColorFool (CF)~\cite{shamsabadi2020colorfool}.  CFX searches for the adversarial example that satisfies two conditions, while CF only considers the activity prediction. The $S_r$ is $\sim$ 80\%  when we attack only the activity prediction. When considering also the explanation similarity, as in $S1$, CF reaches $S_r$ of 37.23 \%, whereas CFX reaches $S_r$ of 64.62 \% (Eq.~\ref{eqn:sr}). 
Similarly, when using  FL-s to attack only the activity, we observed that $\sim$ 66\% of images with different activity have also different explanations ($Q_{\hat{T}}<0.85$).

\newcommand{\markertwo}{\raisebox{0.5pt}{\tikz{\node[draw,scale=0.75,regular polygon, regular polygon sides=4,fill={plotblue}](){};}}}

\newcommand{\markerthree}{\raisebox{0.5pt}{\tikz{\node[draw,scale=0.75,regular polygon, regular polygon sides=4,fill={plotorange}](){};}}}

\section{Conclusion}
\label{sec:conclusion}

We presented a black-box attack on a self-rationalizing multimodal explanation system and evaluated the robustness of its prediction-explanation mechanism under two scenarios: changing the activity prediction while keeping the textual explanations similar and preserving the activity prediction
while modifying the textual explanation. The adversarial examples are generated through semantic colorization or through image filtering. We showed that the prediction-explanation mechanism is vulnerable to black-box attacks that 
use only the final output of the target model. As future work, we will conduct a subjective evaluation of the adversarial examples to inform the attention mechanism.  The  proposed approach could be used to develop model-agnostic evaluation metrics to enable comparative and fair assessment of the faithfulness of different vision-language explanation systems.

{\small
\bibliographystyle{ieee_fullname}
\bibliography{iccv2023AuthorKit/bibliography_updated_full_name}
}

\end{document}